# An Empirical Study of Sections in Classifying Disease Outbreak Reports

**Son Doan[1], Mike Conway[2] and Nigel Collier[2]**

[1]Department of Biomedical Informatics, Vanderbilt University Medical Center, Nashville, TN 37203, US.

[2]National Institute of Informatics, 2-1-2 Hitotsubashi, Chiyoda-ku, Tokyo, Japan

**Abstract** Identifying articles that relate to infectious diseases is a necessary step for any automatic bio-surveillance system that monitors news articles from the Internet. Unlike scientific articles which are available in a strongly structured form, news articles are usually loosely structured. In this chapter, we investigate the importance of each section and the effect of section weighting on performance of text classification. The experimental results show that (1) classification models using the headline and leading sentence achieve a high performance in terms of F-score compared to other parts of the article; (2) all section with bag-of-word representation (full text) achieves the highest recall; and (3) section weighting information can help to improve accuracy.

## 1 Introduction

In infectious disease surveillance systems such as the Global Public Health Intelligence Network (GPHIN) system [14] and ProMed-Mail [7], the detection and tracking of outbreaks using the Internet has been proven to be a key source of information for public health workers, clinicians, and researchers interested in communicable diseases.

For any automatic bio-surveillance system the identification and classification of articles that relate to infectious diseases is a necessary first step in monitoring news articles from the Internet. This reduces the load on later processing steps that often involve more knowledge intensive methods.

In practice though there are a large number of news articles whose main subject is related to diseases but which should not necessarily be notified to users together with a relatively small number of high priority articles that experts should be actively alerted to. Alerting criteria in our approach broadly follow the World Health Organization (WHO) guidelines and include news related to newly emerging diseases, the spread of diseases across international borders, the deliberate release of a human or engineered pathogen, etc.

The use of conventional approaches in the classification process, i.e., bag-of-word, inevitably fails to resolve many subtle ambiguities, for example semantic class ambiguities in polysemous words like "virus", "fever", "outbreak", and "control" which all exhibit a variety of senses depending on context. These different senses appear with relatively high frequency in press news reports especially in headlines where context space is limited and creative use of language is sometimes employed to catch attention. A further challenge is that diseases can be denoted by many variant forms. Therefore we consider that the use of advanced natural language processing (NLP) techniques like named entity recognition (NER) and anaphora resolution are needed in order to achieve high classification accuracy.

In recent years, there have been many studies on text classification in general [17, 23], or on semi-structured texts [10], and XML classification [25]. Other research has investigated the contribution of linguistic information in the form of synonyms, syntax, etc. in text representation [2, 4, 6] or feature selection [16].

In this chapter we focus on an empirical study of canonical sections in news articles. The sections of news are based on the thematic superstructure approach by van Dijk [20]. In the experiments reported in this chapter we adopt a simple approach by dividing sections in news reports into four canonical sections, namely, Head, Lead, Content, and Comment; and explore the importance of each section. The main contribution of this chapter is firstly to provide empirical evidence about the importance of each section within news articles related to the relevancy criteria of our domain. This study also suggests that the use of all section with bag-of-word representation in practical systems is very important because it achieves very high recall which is a very important measure in practice.

The rest of this chapter is organized as follows: in Section 2, we provide a brief overview of related work on the importance of sections in topic classification. Next, we introduce an approach based on van Dijk in Section 3. In the Section 4, we outline the BioCaster schema for the annotation of terms in biomedical text; Section 5 provides details of the method, experimental results and analysis on the gold standard corpus. Finally we draw some conclusions in Section 6.

## 2 Related Work

The first question raised relates to how we can identify sections in a news article. So far, there has been surprisingly little work on identifying sections in this genre. Most previous work we are aware of has dealt with scientific articles with clear document structures. These structures usually are Title, Abstract, Introduction, Material and Methods, Results, and Discussion. [13] analyzed the structure of articles using zone identification, in which a general schema for identifying zones within biological articles was proposed. This schema consisted of Background, Problem setting, Outline, Textual, Method and Result.

Based on available structures, there are several works on studying the effects of sections within scientific papers, Sinclair and Webber [19] studied classification of scientific article into GeneOntology codes. They explored several document representations like bag-of-words, bag-of-nouns, bag-of-stemmed nouns on two classifiers (maximum entropy and naïve Bayes) and found that naïve Bayes achieved the best results with bag-of-words document representation. They reported that both Title and Abstract achieved better result than other sections. Additionally, "all section" (that is, full-text) achieved the highest recall measure.

Related to document representation, Yildiz et al. [24] reported that using a combination representation of bag-of-phrases and bag-of-words can improve the text classification in MEDLINE articles, however the improvement is small (0.03% F-score).

Shah et al. [18] studied the information distribution within biological articles and showed that Abstracts provided the highest keyword density, but other sections might be better sources for the extraction of biologically relevant data.

Schuemie et al. [15] analyzed further the distribution of information in biomedical full-text articles. They gave two criteria, information density and information coverage, to measure the distribution of information within text. They found that Abstracts contain the highest information density. Moreover, 30% gene symbols in the Abstract were accompanied by their names, compared to 18% in the full text.

A recent study on section weighting by Hakenberg et al. [5] investigated several weight methods for sections and found that by setting putting the greatest weight on Abstract and Introduction sections the system achieved the best performance in terms of F-score.

In this work we are working with news articles which have relatively loose structures. Based on the work of van Dijk, we simply divide a news article into 4 canonical sections: Head, Lead, Content, and Comment. For document representation, we use the methodology described in [19] with a bag-of-words representation. For information density analysis, we used information density as described in [15] for analyzing concepts in Summary related to the main topic. Finally, for section weighting, we used the best results reported in [5] by setting the Summary weight to 3. Additionally, we range the weight to 5, 10 to show the effect of section weighting.

## 3 Simply Thematic Structures in Epidemic News

Identifying structures inside news articles is non-trivial. In a scientific article we can easily recognize separate sections such as Title, Abstract, Introduction, Method and Conclusion by their titles. News is different to scientific article in two respects: length and structure. Length is typically less than 600 words from major news agencies and structure is often flexible depending on the context of the story and the reporter.

We firstly consider a thematic approach to news proposed by van Dijk [20]. In his work, he proposed a thematic schema of news, namely Headline, Lead (together forming the Summary), Main Events, Context, History (together forming the Background), Verbal Reactions, and Comments. The Headline, according to van Dijk, "has a very specific thematic function: it usually expresses the most important topic of the news item". The next section is Lead, which often opens a news article. Headline and Lead both directly express the highest level macro-propositions of the news. Together, then, they function as a summary for the news article, thus they form the Summary section. The next section is the Background which includes several sub-categories: Main Events, Context, History and reflects the content of the news. The following section is Verbal Reactions which is defined as quotations and reflects opinions of peoples related to the news. Finally Comment section often contains conclusions, expectations, speculations, and other information about the news. From these descriptions of the various sections that constitute a typical news article, two obvious questions arise: How to identify these sections within news article and how important is each section to the main topical relevance of news articles? As we discussed earlier, the structures of a news article is quite loose, thus it seems very difficult to apply van Dijk's sections directly to news articles. In discussions with experts in public health, we realized that in many (thought not all) cases the main topic and its relevance to infectious disease can be detected when they scanned the headline and the first sentence of a news article rather than the whole text. By incorporating this idea to the van Dijk approach, we assume a simplified news structures with four main sections as follows:

1. Headline: Title of a news article, on the top of a news.
2. Lead: The first sentence in a news article following the Headline. Headline and Lead form the Summary section.
3. Content: Background and Verbal of a news article. It is the section following the Lead.
4. Comment: The last sentence of a news article.

The structure of a news article can be schematically represented as follows:
<HEADLINE> ... </HEADLINE>
<TIME> ...</TIME> (optional)
<LOCATION>...</LOCATION> (optional)
<SENTENCE 1> ... </SENTENCE 1>
<SENTENCE 2> ... </SENTENCE 2>
...
<SENTENCE n-1> ... </SENTENCE n-1>
<SENTENCE n> ... </SENTENCE n>

For example:
<HEADLINE> Cholera in Angola - update </HEADLINE>
<TIME> 21 June 2006 <TIME>
<SENTENCE 1> As of 19 June 2006, Angola has reported a total of 46758
cases including 1893 deaths with an overall (case fatality rate,CFR 4.0%).
</SENTENCE 1>
...
<SENTENCE 7> WHO is sending Interagency Diarrhoeal Disease Kits
to the most affected provinces and continues to support
the Ministry of Health in its surveillance, water and sanitation, social
mobilization and logistics activities.
</SENTENCE 7>.

Here, the Headline is "Cholera in Angola - update", Lead is "21 June 2006. As of 19 June 2006, Angola has reported a total of 46 758 cases including 1893 deaths with an overall (case fatality rate, CFR 4.0%)". We assume that it includes information about time and organization that is of central importance to the news story. Content is the text between <SENTENCE 2> and </SENTENCE 6>. Comment is the text between <SENTENCE 7> and </SENTENCE 7>. The Summary section is the text inside <HEADLINE> and </SENTENCE 1>.

## 4 Data Set

In addition to section headings we wanted to explore the use of terminology and its classes in our classification models. Below we present a brief summary of the schema and then follow this with a description of the data set used in our experiments.

## 4.1 BioCaster Annotation Schema

The BioCaster annotation schema is a component of the BioCaster text mining project. This schema has been developed for annotating important concepts that reflect information about infectious diseases. These key concepts are classified as Type and Role in which Type is identified using Name Entity Recognition (NER) and Roles are associated as attributes to the Name Entities (NEs). In total there are 18 NEs denoted by convention in upper case. These include PERSON, LOCATION, ORGANIZATION, TIME, DISEASE, CONDITION, OUTBREAK, VIRUS, ANATOMY, PRODUCT, NONHUMAN, DNA, RNA, PROTEIN, CONTROL, BACTERIA, CHEMICAL and SYMPTOM. Of them, PERSON has the attribute case, NONHUMAN and ANATOMY have the attribute transmission, and CHEMICAL has the attribute therapeutic. They are marked up into text in XML format as follows,

<NAME cl="Named Entity" attribute1="value1" attribute2="value2" ... </NAME>,
where "Named Entity" is one of the names for the 18 BioCaster NEs and attribute1, attribute2, ... are the names of the NE's attributes, "value1", "value2", ... are values corresponding to attributes. Further details of the annotation guidelines are discussed in [9].

For example,
<NAME cl="DISEASE">Cholera</NAME> in <NAME cl="LOCATION"> Angola </NAME> - update <NAME cl="TIME">21 June 2006 </NAME> As of <NAME cl="TIME"> 19 June 2006</NAME>, <NAME cl="LOCATION">Angola</NAME> has reported a total of <NAME cl="PERSON" case ="true" number="many">46 758 cases</NAME> including <NAME cl="PERSON" case="true" number="many"> 1893 deaths</NAME> with an overall (case fatality rate, CFR 4.0%). ...
<NAME cl = "ORGANIZATION">WHO</NAME> is sending Interagency Diarrhoeal Disease Kits to the most affected provinces and continues to support the Ministry of Health in its <NAME cl = "CONTROL"> surveillance </NAME>, water and sanitation, social mobilization and logistics activities.

Information about important concepts obtained from annotations can serve as clues for recognizing four sections inside news articles defined in Section 3.

## 4.2 BioCaster Gold Standard Corpus

The BioCaster gold standard corpus was collected from Internet news and manually annotated by two doctoral students. The annotation of a news article proceeded as follows. Firstly, NEs are annotated following the BioCaster schema and guidelines. Secondly, each annotated article is manually assigned into one of four relevancy categories: alert, publish, check, and reject. The assignment is based on guidelines that we made following discussions with epidemiologists and a survey of World Health Organization (WHO) reports [21]. These categories are currently being used operationally by the GPHIN system which is used by the WHO and other public health agencies. Where there were major differences of opinion in NE annotation or relevancy assignment between the two annotators, we consulted a public health expert in order to decide the most appropriate assignment. Finally we had a total of 1000 articles that were fully annotated.

We grouped the 1000 articles into 2 categories: reject and relevant. The reject category corresponds simply to articles with label reject while the relevant category includes articles with labels alert, publish, and check. We conflated the alert, publish and check categories because we hypothesized that distinguishing between non-reject (relevant) categories would require higher level semantic knowledge such as pathogen infectivity and previous occurrence history which is the job of the text mining system and the end user. Finally we had a total of 650 news articles belong to the reject category and 350 news articles belong to the relevant category.

## 5 Experiments

### *5.1 Method*

We used the BioCaster gold standard corpus to investigate the effect of canonical sections on performance of classification. The experimental process was as follows. We randomly divided the data set into 10 parts. Each part has 35 articles belonging to the relevant category and 65 articles belonging to the reject category. Then, we implemented 10-fold cross validation: 9 parts for training and the remaining part for testing. For training sets we extracted canonical sections as features to build a classifier. The remaining part was used for testing. The classifier we used in this chapter is the standard naïve Bayes [12], maximum entropy (baseline maxent) and Support Vector Machine (SVM) classifiers. In the pre-processing we did not use either a stop list or word stemming. The experiments were implemented in Linux OS, using the Bow toolkit [11] for naïve Bayes and maximum entropy; and $SVM_{light}$ [8] for SVM.

In document representation, we used bag-of-words representation for naïve Bayes which was reported in [19] that it offers the best result compared to other methods like bag-of-nouns, bag-of-stems, bag-of-stemmed nouns. We used term frequency as term weight and linear kernel in SVM because they were reported as the best results compared to other weighting methods [1]. For section weighting we used the same section weighting methods described in [5] in which both Abstract and Introduction were set to 3 while the remaining sections were set to 1. In addition, we set the weights of both Headline and Lead sections to 3, 5, and 10 and the remaining sections to 1.

The features for training data are grouped as follows:

1. The baseline

– Baseline: Using full-text only as the baseline method.

2. Section features

– Head: Only the Head section is used as features, i.e., all text, NEs and NE's attributes in the Headline section.

– Lead: Only the Lead section in headline is used as features, i.e., all text, NEs and NE's attributes in the Lead section.

– Content: Only the Content section is used as features, i.e., all text, NEs and NE's attributes in the Content section.

– Comment: Only the Comment section is used as features, i.e., all text, NEs and NE's attributes in the Comment section.

– All sections: All Sections are used as features, i.e., all text, NEs and NE's attributes in the whole news article. These features are reported that they achieved the best result in annotated text classification in [3].

3. Summary features

– Summary text: Only full-text in summary section are used as features, i.e., only full-text in the Summary section.

– Summary NEs: Only NEs in summary section are used as features, i.e., only NEs and their attributes in the Summary section.

– Summary text + NEs: Both text and NEs are used as features, i.e., all full-text, NEs and their attributes in the Summary section.

4. Features of section weights: Both text and NEs are used as features, i.e., all full-text, NEs and their attributes in the whole news article. Of them, the Summary section is weighted as follows:

– Summary text + NEs 3: The Summary section are set to 3, all remainings are set to 1.

– Summary text + NEs 5: The Summary section are set to 5, all remainings are set to 1.

– Summary text + NEs 10: The Summary section are set to 10, all remainings are set to 1.

In order to investigate some linguistic properties related to the main topic of news, we use the concept of information density described in [15] in which we define the information density of each NE in Summary as its frequency in the Headline section over the whole corpus. We will describe this in detail in Section 5.3.

|  | YES is correct | NO is correct |
|---|---|---|
| Assigned YES | a | b |
| Assigned NO | c | d |

**Table 1**. A contingency table.

## 5.2 Performance measures

In our experiments we use two performance measures, standard Precision/Recall and accuracy. Both are calculated based on the two-way contingency table shown in Table 1. There a is the frequency of assigned and correct cases, b is the assigned and incorrect cases, c is the none assigned but incorrect cases, and d counts the none assigned and correct cases [22]. Then,
Precision = a/ (a + b), and Recall = a/(a + c).
Accuracy and F-score are defined as accuracy=(a + d)/(a + b + c + d),
F-score =2 × Precision × Recall / (Precision + Recall).

## 5.3 Results and Discussions

The results of experiments for each algorithm are shown in Table 2. We next discuss the effectiveness of sections based on evaluations of algorithms, sections, NEs and section weighting. Algorithm Evaluation Of the three algorithms, naïve Bayes achieved the best F-score with 85.60%, while SVM achieved 81.63% and MaxEnt achieved 84.15%, respectively. These confirmed the results in [19] for scientific papers. Hereafter, we will focus on the improvement in naïve Bayes.

**Sections Evaluation**

We now investigate the effectiveness of each section in a news article. Interestingly we saw the Lead section achieves the best F-score with 85.6% using naïve Bayes. It indicates that the first sentence in a news article play a very important role in deciding whether the news article should be alerted or not. This reflects the public health experts' comment mentioned in Section 3 that they can often judge the relevancy of a news article by scanning the first sentence. The contribution of each section to accuracy can be ranked in the following order: Lead > Content > All Section > Headline for naïve Bayes. However it is not conclusive because we observed there are not consitent in SVM and MaxEnt methods in Table 2. We also noticed that in both naïve Bayese and SVM the highest Recall achieved is 98.29 when using baseline method which is all sections with bag-of-word representation. This suggests that the bag-of-word representation method still plays very important in practice not only due to its simplicity but its performance.

| FEATURES | Naïve Bayes | SVM | MaxEnt |
|---|---|---|---|
| Baseline | 81.60/65.90/98.29 78.90 | 68.70/53.54/98.29 69.32 | 81.70/86.15/56.86 68.51 |
| Headline | 84.00/82.31/69.14 75.15 | 84.50/75.47/84.57 79.76 | 79.70/80.48/55.43 65.65 |
| Lead | 88.90/78.38/94.29 85.60 | 77.20/61.62/98.86 75.92 | 77.70/86.71/42.86 57.36 |
| Content | 87.90/75.28/97.43 84.93 | 85.80/81.46/78.29 79.84 | 88.70/83.57/84.29 83.93 |
| Comments | 85.60/71.55/97.71 82.61 | 76.40/60.97/100.0 75.75 | 85.20/82.58/73.14 77.57 |
| All sections | 88.10/80.00/88.10 83.85 | 86.40/84.36/75.71 79.80 | 73.10/90.10/26.00 40.35 |
| Summary text only | 54.60/43.50/99.43 60.52 | 40.00/36.85/100.0 53.86 | 59.10/45.99/96.57 62.31 |
| Summary NEs only | 82.70/71.64/83.71 77.21 | 82.30/67.88/96.57 79.72 | 81.10/74.47/69.43 72.00 |
| Summary text + NEs | 88.40/78.40/92.29 84.78 | 84.50/71.47/95.14 81.63 | 84.70/86.35/66.86 75.37 |
| Summary text + NEs 3 | 88.20/76.01/96.86 85.18 | 81.20/81.92/60.86 69.84 | 89.00/84.88/83.43 84.15 |
| Summary text + NEs 5 | 88.30/76.66/95.71 85.13 | 75.00/80.66/37.14 50.87 | 88.00/84.34/80.29 82.27 |
| Summary text + NEs 10 | 88.50/77.91/93.71 85.08 | 68.20/75.40/12.29 21.13 | 87.70/86.03/77.43 81.50 |

**Table 2**. Experimental results for 3 algorithms, the number in each column indicates accuracy, Precision/Recall and F-score measures, respectively.

**NE Evaluation**

Now we consider the effect of NEs in the "top sections" (Summary NEs only) on classification accuracy. We can see that using NEs in Summary have comparable F-score to Headline for three algorithms (77.21% vs. 75.15% for naïve Bayes, 79.72% vs. 79.76% for SVM, and 72.0% vs. 65.65.% for maximum entropy).

Looking at the distribution of NEs, we recognize the distinctive tendencies between NEs. Ignoring general NEs like PERSON, LOCATION, ORGANIZATION, TIME, we observed that there are big differences in NEs: NEs that tend to be the relevant category (relevant NEs) which their frequency in the relevant category is much higher than the reject category. Also, NEs tend to be the reject category (irrelevant NEs) which their frequency in the relevant category is much lower than the reject category. The list of relevant NEs are DISEASE, CONDITION, OUTBREAK, VIRUS, SYMPTOM and irrelevant NEs are NONHUMAN, CONTROL, ANATOMY, PRODUCT, BACTERIA, CHEMICAL, PROTEIN, DNA, RNA.

Through this analysis we can see some interesting properties related to linguistic properties of the text. Firstly, the relevant category contains articles about the name of infectious diseases (DISEASE, OUTBREAK), or situations of diseases like conditions, symptoms, or virus cases the disease; secondly, the relevant category contains articles about entities which are not directly related to diseases like proteins, DNA, RNA, drug products. Furthermore, the existence of NONHUMAN and BACTERIA in irrelevant category also indicated about information of "species": documents which are not directly related to people should not belong to the relevant category.

**Section Weighting Evaluation**

Now we consider the effect of section weighting on performance. We can easily see that the top section, i.e., Summary, achieved high performance. The F-score of both algorithms naïve Bayes and maximum entropy are higher than the baseline when using section information. In practice, we observed that SVM does not work well in section weighting: when the weights are set to greater values, the recall tends to drop significantly. We also saw that using section weighting achieved more stable and higher results for both naïve Bayes and maximum entropy. The results suggested that Summary section achieved the highest performance when it is weighted to 3.

## 6 Conclusions

This chapter has focused on analyzing the contribution of section information to the automatic classification of news articles related to disease outbreaks. The experimental results indicated that:
1. Top sections within news, i.e., Headline and Lead play an important role in deciding the main topic of the news.
2. All section with bag-of-word representation (full text) achieves the highest recall.
3. Using section weighting can improve performance of text classification.

In the future we will investigate automatic algorithms for identifying main sections within news articles based on their contributions to the main topic.

References

1. A.R. Aronson, O. Bodenreider, D. Demner-Fushman, K.W. Fung, V.K. Lee, J.G. Mork, A.Neveol, L. Peters, and W.J. Rogers. From indexing the biomedical literature to coding clinical text: Experience with MTI and machine learning approaches. In Proceeding of ACL workshop on BioNLP 2007: Biological, Translation and clinical language processing, pages 105–112, 2007.
2. S. Bloehdorn and A. Hotho. Boosting for text classification with semantic features. In Proc. of the Workshop on Mining for and from the Semantic Web at the 10th ACM SIGKDD 2004, pages 70–87, 2004.
3. S. Doan, A. Kawazoe, and N. Collier. The role of roles in classifying annotated biomedical text. In Proceeding of ACL workshop on BioNLP 2007: Biological, Translation and clinical language processing, pages 17–24, Prague, Czech, 2007.
4. J. Frürnkranz, T. Mitchell, and E. Riloff. A case study in using linguistic phrases for text categorization on the WWW. In Working Notes of the AAAI/ICML Workshop on Learning for Text Categorization, pages 5–13, 1998.
5. J. Hakenberg, J. Rutsch, and U. Leser. Tuning text classification for hereditary diseases with section weighting. In Proc. of the first Int'l Symposium on Semantic Mining in Biomedicine (SMBM), pages 34–37, 2005.
6. A. Hotho, S. Staab, and G. Stumme. WordNet improves text document clustering. In Proc. of the SIGIR 2003 Semantic Web Workshop, 2003, 2003.
7. International Society for Infectious Diseases. ProMed Mail, 2001. http://www.promedmail.org.
8. T. Joachims. Making large-scale SVM Learning Practical . In B. Sch¨olkopf, C. Burges, and A. Smola, editors, Advances in Kernel Methods - Support VectorLearning. MIT press, 1999.
9. A. Kawazoe, L. Jin, M. Shigematsu, R. Barrero, K. Taniguchi, and N. Collier.The development of a schema for the annotation of terms in the BioCaster diseasedetection/tracking system. In Proceedings of the International Workshop on Biomedical Ontology in Action (KR-MED 2006), pages 77–85, 2006.
10. T. Kudo and Y. Matsumoto. A boosting algorithm for classification of semistructured text. In Proceedings of the 2004 Conference on Empirical Methods in NLP, pages 301–308, 2004.
11. A.K. McCallum. Bow: A toolkit for statistical language modeling, text retrieval, classification and clustering, 1996. http://www.cs.cmu.edu/_mccallum/bow.
12. T.M. Mitchell. Machine Learning. Mc Graw Hill, 1997.
13. Y. Mizuta and N. Collier. Zone identification in biology articles as a basis for information extraction. In Proc. of Natural Language Processing in Biomedicine and its Applications (JNLPBA) 2004, pages 29–35, 2004.
14. Public Health Agency of Canada. Global Public Heath Intelligence Network (GPHIN), 2004. http://www.gphin.org.
15. M.J. Schuemie, M. Weeber, B.J.A Schjivenaars, E.M. van Mulligen, C.C. van der Eijik, R. Jellier, B. Mons, and J.A. Kors. Distribution of information in biomedical abstracts and fulltext publications. Bioinformatics, 20:2597–2604, 2004.
16. S. Scott and S. Matwin. Feature engineering for text classification. In Proc. of International Conference on Machine Learning 1999, pages 379–388, 1999.
17. F. Sebastiani. Machine learning in automated text categorization. ACM computing survey, 34(1):1–47, 2002.
18. P.K. Shah, C. Perez-Iratxeta, P. Bork, and M.A. Andrade. Information extraction from fulltext scientific articles: where are the keywords ? BMC Bioinformatics, 4(20), 2003.

19. G. Sinclair and B. Webber. Classification from fulltext: A comparison of canonical sections of scientific papers. In Proc. of Natural Language Processing in Biomedicine and its Applications (JNLPBA) 2004, pages 66–69, 2004.
20. T.A. van Dijk. Structures of news in the press. In Discourse and Communication, pages 69–93. Berlin: De Gruyter, 1985.
21. World Health Organization. ICD10, International Statistical Classification of Diseases and Related Health Problems, Tenth Revision, 2004.
22. Y. Yang. An evaluation of statistical approaches to text categorization. Information Retrieval Journal, 1:69–90, 1999.
23. Y. Yang and X. Liu. A re-examination of text categorization methods. In Proc. of 22th SIGIR, ACM Intl. Conf. on Research and Development in Information Retrieval, pages 42–49, 1999.
24. M. Yetisgen-Yildiz and W. Pratt. The effect of feature representation on MEDLINE document classification. In AMIA Annu Symp Proc., pages 849–853, 2005.
25. M.J. Zaki and C.C. Aggarwal. XRules: an effective structural classifier for XML data. In Proceedings of the ninth ACM SIGKDD international conference, 2003, pages 316–325, 2003.